\documentclass[10pt,twocolumn,letterpaper]{article}

\usepackage{graphicx}
\usepackage{subcaption}

\usepackage{amsfonts}
\usepackage{amsmath}
\RequirePackage{booktabs}
\usepackage{xspace}
\usepackage{hyperref}  % 必须先加载
\usepackage{cleveref}

\newcommand{\relu}{ReLU}
\newcommand{\clamp}{clamp}
\newcommand{\CE}{CE}
\newcommand{\Skew}{Skew}
\newcommand{\VSM}{VSM}
\newcommand{\SoftVSM}{Soft\text{-}VSM}
\newcommand{\Oracle}{Oracle}

\newcommand{\ind}{\mathbb{1}}
\newcommand{\shat}{\hat{s}_j}
\newcommand{\vhat}{\hat{v}_j}
\newcommand{\Ts}{T_s}
\newcommand{\Tv}{T_v}
\newcommand{\etal}{\emph{et al.}\xspace}

\makeatletter
\DeclareRobustCommand\onedot{\futurelet\@let@token\@onedot}
\def\@onedot{\ifx\@let@token.\else.\null\fi\xspace}

 \def\vs{\emph{vs}\onedot}

\def\etal{\emph{et al}\onedot}
\makeatother

\title{A Threshold-Free Visual-Semantic Matching Metric for Evaluating Consistency in Subject-Driven Image Generation}

\author{Bowei Shan, Heng Yin\\
School of Information Engineering, Chang'an University, Xi'an, China
}

\begin{document}
\maketitle

%-------------------------------------------------------------------------
% Two-column teaser: Figure 1 of the source paper (framework overview)
%-------------------------------------------------------------------------

%-------------------------------------------------------------------------
\begin{abstract}
We propose a Threshold-Free Visual-Sematic Matching (TF-VSM) metric, to quantitatively evaluate localized visual consistency across varying poses and environments in subject-driven text-to-image generation. TF-VSM is a fully continuous, threshold-free visual consistency metric paired with an extended feature-disentanglement retraining schedule for dual-branch ResNet feature-aggregation networks. Existing metrics suffer from point-alignment dropouts, threshold sensitivity, and score instability. To address them, our TF-VSM, rather than enforcing rigid binary cutoffs ($T_s$, $T_v$), weights every object point by a non-linear semantic match confidence $w_j = ReLU(\hat s_j)^2$ to yield a closed-form confidence-weighted score without any hand-tuned hyperparameters. Experimental results show that, TF-VSM attains Pearson 0.638 and Spearman 0.674 oracle correlation, improving over the pretrained VSM baseline (0.454/0.589) by +0.171/ + 0.116 after retraining, and drastically outperforming CLIP (-0.052/ -0.004), DINO  (0.087/0.121), and VLMs (0.096/0.003). Six diagnostic ablations confirm that smooth confidence weighting is the right mechanism, that hard semantic gating is brittle (dropping to 0.533 Pearson at $T_s$=0.8), and that the threshold-free metric is essentially invariant to the training time loss balance $\alpha$. Code is available at: \url{https://github.com/bwshan/TF-VSM}.
\end{abstract}

%=========================================================================
\section{Introduction}
\label{sec:intro}

Subject-driven text-to-image generation has advanced rapidly, enabling users to personalize diffusion models to synthesize specific target subjects across diverse poses, lighting, and background environments~\cite{ruiz2022,gal2022,kumari2022,wei2023,hao2023,li2023,tan2024,zhang2025b}. A central requirement in evaluating these models is assessing \emph{visual consistency}---determining whether the visual identity and subtle textural details of a subject are faithfully preserved across reference and generated image pairs~\cite{tewel2024,avrahami2023}. An ideal metric must remain invariant to variations in pose, background, scale, and perspective while remaining sensitive to fine-grained visual inconsistencies such as altered surface patterns, modified logos, or subtle structural artifacts. Despite its importance, quantifying visual consistency remains an open challenge. Prior evaluation protocols predominantly rely on two paradigms:

\begin{enumerate}
\item \textbf{Global feature similarity \& vision-language models.} Standard frameworks compute distances between global image embeddings such as CLIP-I~\cite{ruiz2022,gal2022} or DINO-I~\cite{caron2021,oquab2023}. Because these representations pool features over spatial dimensions, they conflate high-level identity with layout and are blind to localized inconsistencies~\cite{abbasi2025,chen2025}. Likewise, VLMs (e.g., GPT-4o, Gemini) and VLM-driven suites~\cite{ku2023a,peng2024,hua2025} offer impressionistic judgments that struggle to track fine attributes and aggregate local visual evidence across images~\cite{lin2024,das2026}.

\item \textbf{Threshold-based point alignment \& feature disentanglement.} To overcome spatial pooling, recent approaches such as Visual-Semantic Matching (VSM)~\cite{eldesokey2025} extract internal features from pre-trained diffusion backbones~\cite{rombach2021,stracke2024} and train dual aggregation networks to disentangle semantic structure from visual appearance. Standard VSM, however, computes consistency by applying hand-tuned binary thresholds $(\Ts,\Tv)$ to filter semantic keypoints and to binarize visual similarities. As decision-boundary studies show~\cite{aksoy2025,moslemi2024}, hard thresholding creates step-discontinuities that cause severe point-alignment dropouts and score instability whenever valid correspondences fall slightly below arbitrary cutoffs.
\end{enumerate}

\begin{figure*}[t]
\centering
\includegraphics[width=\linewidth]{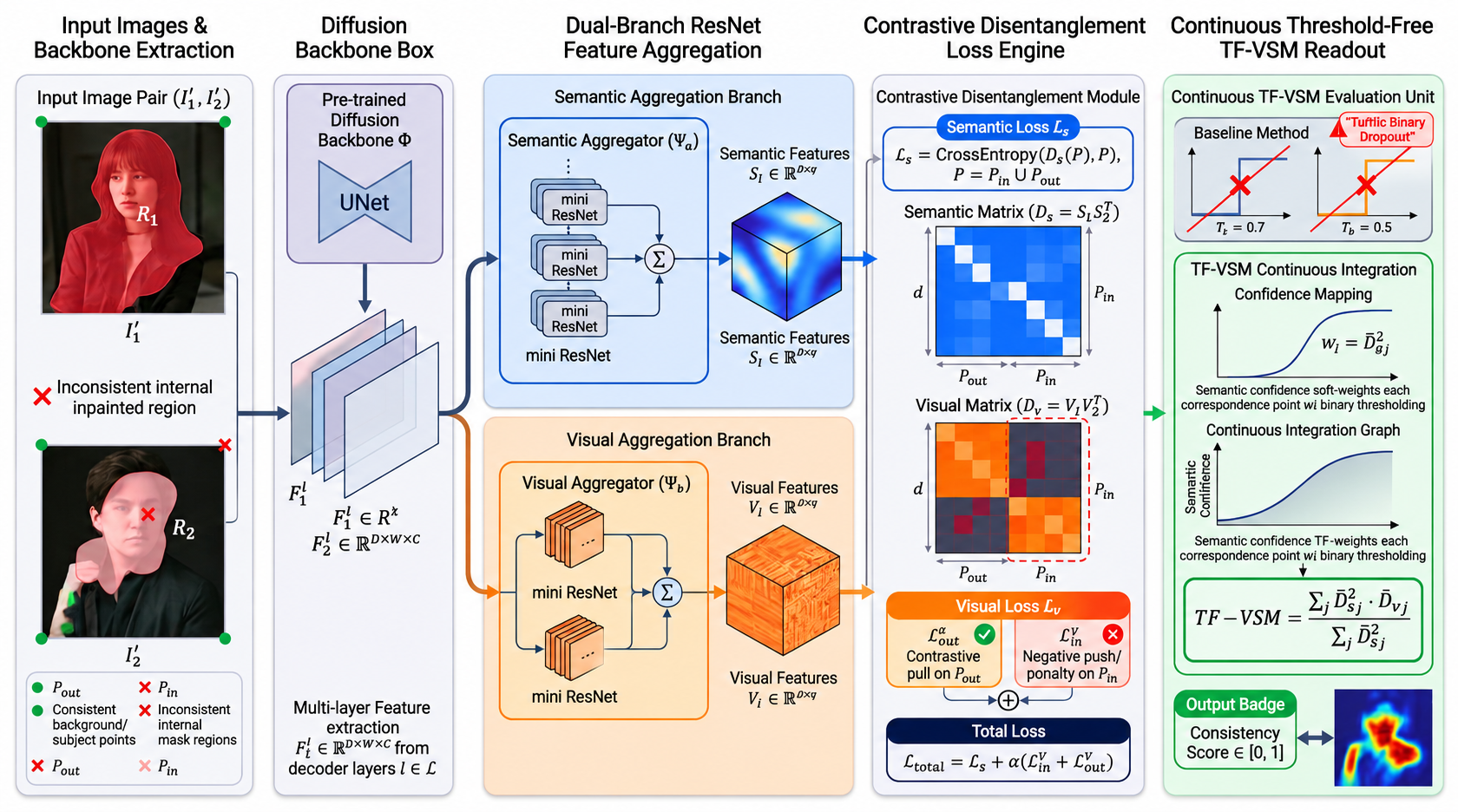}

\caption{\textbf{Overview of the proposed framework.} Given an input pair $(I'_1,I'_2)$ with synthetic inpainted regions $(R_1,R_2)$, multi-layer features are extracted from a frozen diffusion backbone $\Phi$ and processed by dual-branch ResNet aggregation networks $(\Psi_s,\Psi_v)$ into semantic features $S_i$ and visual features $V_i$. A contrastive disentanglement module enforces pose-invariant semantic matching over all correspondences while penalizing visual similarity inside inconsistent regions. The threshold-free TF-VSM unit then reads out a consistency score by soft-weighting visual similarities with squared semantic confidence $w_j=\relu(\shat)^2$, avoiding brittle binary cutoffs $(\Ts,\Tv)$ and producing a global score alongside a diagnostic heatmap.}

\label{fig:overview}
\end{figure*}

To address these limitations, we introduce \textbf{TF-VSM} (Threshold-Free Visual-Semantic Matching), a continuous, fully threshold-free visual semantic consistency metric. Rather than gating correspondences with a rigid cutoff $\Ts$, TF-VSM assigns every object point a smooth semantic-match weight $w_j=\relu(\shat)^2$, so each point contributes according to its alignment confidence. It further eliminates the visual threshold $\Tv$ by integrating the visual indicator over all consistency levels $r\in[0,1]$, which reduces in closed form to a confidence-weighted soft visual similarity with no hand-tuned cutoffs. In tandem, we retrain the dual feature-aggregation networks on the Subjects200k-derived dataset~\cite{tan2024,eldesokey2025} with an extended schedule, identifying an optimal convergence checkpoint (epoch 8) that markedly improves feature disentanglement. Through controlled evaluation on a 100-pair test set with ground-truth inpainting oracles, TF-VSM achieves Pearson $0.638$ and Spearman $0.674$ correlation with the oracle, markedly outperforming global CLIP ($-0.052/-0.004$), DINO ($0.087/0.121$), Gemini VLM ($0.096/0.003$), and the original pretrained VSM ($0.454/0.589$). Our contributions are:

\begin{itemize}
\item A fully threshold-free visual-consistency metric that removes both $\Ts$ and $\Tv$ via non-linear confidence weighting and continuous visual integration, yielding a simple closed form.
\item An extended disentanglement retraining recipe that lifts oracle correlation from $0.467/0.558$ to $0.638/0.674$ ($+0.171/+0.116$) over the released model.
\item Six diagnostic ablations isolating the confidence weighting, the semantic/visual readout transforms, the schedule length, the data scale, and the loss balance $\alpha$, showing TF-VSM matches oracle-tuned hard gating while removing its brittle hyperparameter dependence.
\item A leakage-free evaluation protocol (disjoint selection/test splits) that independently recovers the epoch-8 checkpoint and confirms the gain on a strictly held-out split, together with foreground-masked baselines showing that background removal explains almost none of the gap to global metrics.
\end{itemize}

%=========================================================================
\section{Related Work}
\label{sec:related}

We contextualize our work across three paradigms: global feature similarity and VLM evaluation, dense semantic correspondence via diffusion features, and threshold-based consistency metrics.

\noindent\textbf{Global feature similarity \& vision-language evaluation.}
Subject-driven personalization models~\cite{ruiz2022,gal2022,kumari2022,wei2023,hao2023,li2023,ma2023} synthesize user-specified subjects across diverse contexts, and their evaluation requires standardized benchmarks~\cite{ku2023b,huang2023a,slobodkin2025}. Fidelity is historically quantified via global embedding distances using CLIP-I~\cite{ruiz2022} or self-supervised backbones such as DINO/DINOv2~\cite{caron2021,oquab2023,wysoczanska2023,tschannen2025}. However, global pooling makes these metrics insensitive to localized textural discrepancies~\cite{abbasi2025,fu2023,napolitano2026}, motivating localized perceptual variants such as matched-points LPIPS~\cite{chen2024}, human-aligned customized LPIPS~\cite{trinh2026}, and foreground-masked patch matching~\cite{bartkowiak2026}. In parallel, VLM-driven protocols such as DreamBench++~\cite{peng2024}, VIEScore~\cite{ku2023a}, VQAScore~\cite{lin2024}, MMIG-Bench~\cite{hua2025}, and FineMatch~\cite{hua2024} assess image--text alignment, yet exhibit spatial-reasoning bottlenecks~\cite{chen2025}, cross-image aggregation failures~\cite{das2026}, and reliance on text-mediated pathways~\cite{salazar2026,oh2026}. Recent suites such as USO-Bench~\cite{wu2025} and RefVNLI~\cite{slobodkin2025} evaluate subject consistency but do not provide dense point-wise disentanglement of appearance from structure.

\noindent\textbf{Dense semantic correspondence via diffusion features.}
Pretrained diffusion backbones~\cite{rombach2021,podell2023} implicitly learn rich spatial-semantic representations. DIFT~\cite{tang2023} and Plug-and-Play~\cite{tumanyan2022} showed that internal UNet features enable zero-shot dense keypoint matching across challenging poses. Subsequent work expanded these capabilities via SD--DINO fusion~\cite{zhang2023a}, Diffusion Hyperfeatures~\cite{luo2023}, geometry-aware modeling~\cite{zhang2023b}, spherical priors~\cite{mariotti2023}, lightweight ViT feature transforms~\cite{suri2024}, prompt-attention optimization~\cite{hedlin2023}, and noise-free extraction via CleanDIFT~\cite{stracke2024}, with distillation~\cite{fundel2024} and conditional matching~\cite{nam2023} improving efficiency. Surveys~\cite{zhang2025a,chong2025,dunkel2026,cuttano2026,huang2023b} note that self-supervised ViT descriptors~\cite{amir2021} and diffusion features achieve strong semantic alignment but intentionally discard appearance variation. Consequently, correspondence-driven consistency methods---ConsiStory~\cite{tewel2024}, Eye-for-an-eye~\cite{go2024}, ReStyle3D~\cite{zhu2025}, The Chosen One~\cite{avrahami2023}, Self-Cross Guidance~\cite{qiu2024}, and Grounding DINO~\cite{liu2023}---focus on driving synthesis rather than explicitly evaluating visual consistency.

\noindent\textbf{Threshold-based consistency metrics \& subject-driven benchmarks.}
Automated generation and evaluation workflows~\cite{cai2024,shin2024,zhang2025b} often leverage synthetic dataset pipelines. Pairing identity-consistent datasets such as Subjects200k~\cite{tan2024} with SAM~\cite{kirillov2023} and SDXL inpainting~\cite{podell2023} yields controlled pairs with localized variations~\cite{eldesokey2025}. However, perceptual metrics such as LPIPS~\cite{zhang2018} and the original VSM~\cite{eldesokey2025} rely on hand-tuned scalar thresholds $(\Ts,\Tv)$; decision-boundary studies~\cite{aksoy2025,moslemi2024} show that single-threshold binarization introduces selection bias where minor cutoff shifts drastically alter rankings. TF-VSM removes these artifacts by replacing binary cutoffs with continuous, confidence-weighted integration.

%=========================================================================
\section{Methodology}
\label{sec:method}

We present the technical framework of TF-VSM: an automated data-generation pipeline producing paired data with known localized inconsistencies (\cref{sec:data}); a dual-ResNet aggregation architecture with a contrastive disentanglement objective (\cref{sec:disentangle}); and the continuous, threshold-free TF-VSM metric (\cref{sec:softvsm}). \Cref{fig:overview} gives an overview.

%-------------------------------------------------------------------------
\subsection{Data Generation Pipeline for Inconsistency Supervision}
\label{sec:data}

To learn to separate semantic structure from visual appearance, we construct a supervision dataset of paired images with precise spatial masks marking consistent and inconsistent regions, starting from identity-consistent pairs $(I_1,I_2)$ in Subjects200k~\cite{tan2024}.

We first isolate the foreground subject in both images with Grounded-SAM~\cite{kirillov2023,liu2023}, yielding binary subject masks $O_1,O_2\in\{0,1\}^{h\times w}$. To establish semantic correspondences, we extract features from the 6th UNet decoder layer of Stable Diffusion 2.1~\cite{rombach2021,stracke2024}, $F^6_1=\Phi(I_1)$ and $F^6_2=\Phi(I_2)$, and form the dense similarity matrix $D=F^6_1 (F^6_2)^\top\in\mathbb{R}^{d\times d}$ with $d=h\times w=48\times 48=2304$. Dense correspondences $C_1,C_2\in\mathbb{R}^{N\times 2}$ are recovered by spatial argmax:
\begin{equation}
\begin{aligned}
C_1 &= \{(x^1_1,y^1_1),\dots,(x^1_N,y^1_N)\},\\
C_2 &= \{(x^2_1,y^2_1),\dots,(x^2_N,y^2_N)\}.
\end{aligned}
\label{eq:corresp}
\end{equation}
To create localized inconsistencies, we select a high-confidence match $k$ and prompt SAM to segment corresponding sub-regions $R_i=\mathrm{SAM}(I_i,(x^i_k,y^i_k))$, returning multiple hierarchical candidates and picking the smallest-footprint mask to avoid segmenting the whole object.

\noindent\textbf{Ambiguity filtering via similarity skewness.}
On flat, textureless regions (plain surfaces, uniform clothing) point $k$ matches diffusely. We measure the skewness of $D[k]$,
\begin{equation}
\Skew(D[k]) = \frac{n}{(n-1)(n-2)}\sum_{i=1}^{n}\left(\frac{D[k]_i-\mu}{\sigma}\right)^3,
\label{eq:skew}
\end{equation}
with $n=|D[k]|$, $\mu=\mathrm{mean}(D[k])$, $\sigma=\mathrm{STD}(D[k])$: high skewness indicates sharp localized matches, low skewness ambiguous ones. We keep samples with $\Skew(D[k])>1.3$.

\noindent\textbf{Inpainting and perceptual validation.}
For regions passing structural checks we crop a padded patch and perform localized SDXL inpainting~\cite{podell2023}, so that new textures are synthesized without global object context. We then validate perceptual change via LPIPS~\cite{zhang2018}, discarding pairs with $\mathrm{LPIPS}\le 0.15$. The resulting dataset contains 4,912 usable training pairs $(I'_1,I'_2)$ with subject masks $(O_1,O_2)$ and inconsistency masks $(R_1,R_2)$.

%-------------------------------------------------------------------------
\subsection{Disentangled Feature Aggregation and Retraining Objective}
\label{sec:disentangle}

Given an inconsistent pair $(I'_1,I'_2)$ with masks $R_1,R_2$, we learn descriptors separating pose-invariant semantic structure from visual appearance. Multi-layer feature maps $F_i=\Phi(I'_i)$ across decoder layers $l\in\mathcal{L}$ are passed through two aggregation networks $\Psi_s$ (semantic) and $\Psi_v$ (visual), each a per-layer mini-ResNet~\cite{eldesokey2025} combined via trainable scalar weights $w^s_l,w^v_l$:
\begin{equation}
S_i = \sum_{l\in\mathcal{L}} w^s_l\,\Psi^l_s(F^l_i),\quad
V_i = \sum_{l\in\mathcal{L}} w^v_l\,\Psi^l_v(F^l_i),\quad
S_i,V_i\in\mathbb{R}^{d\times q},
\label{eq:agg}
\end{equation}
with projected channel dimension $q=384$ and $d=2304$. We form spatial correlation matrices $D^{12}_s=S_1S_2^\top$, $D^{21}_s=S_2S_1^\top$, $D^{12}_v=V_1V_2^\top$, $D^{21}_v=V_2V_1^\top$, and partition correspondences into interior ($P^i_{\mathrm{in}}$, inside $R_i$) and exterior ($P^i_{\mathrm{out}}$) subsets:
\begin{equation}
\begin{aligned}
P^i_{\mathrm{in}} &= \{(x^i_j,y^i_j)\in C_i \mid R_i(x^i_j,y^i_j)=1\},\\
P^i_{\mathrm{out}} &= \{(x^i_j,y^i_j)\in C_i \mid R_i(x^i_j,y^i_j)=0\}.
\end{aligned}
\label{eq:part}
\end{equation}
Semantic features are pulled together across all matches via a cross-entropy loss over $P_i=P^i_{\mathrm{in}}\cup P^i_{\mathrm{out}}$, while visual features are pulled together outside and pushed apart inside inconsistent regions:
\begin{equation}
\begin{aligned}
\mathcal{L}_s &= \CE\big(D^{12}_s(P_1),P_2\big),\\
\mathcal{L}^v_{\mathrm{in}} &= \CE\big(-D^{12}_v(P^1_{\mathrm{in}}),P^2_{\mathrm{in}}\big),\\
\mathcal{L}^v_{\mathrm{out}} &= \CE\big(D^{12}_v(P^1_{\mathrm{out}}),P^2_{\mathrm{out}}\big).
\end{aligned}
\label{eq:loss}
\end{equation}
Combining terms symmetrically over both directions gives $\mathcal{L}=\mathcal{L}_s+\alpha(\mathcal{L}^v_{\mathrm{in}}+\mathcal{L}^v_{\mathrm{out}})$, where $\alpha=10$ prioritizes visual learning. We optimize with AdamW ($\eta=5\times10^{-4}$) over an extended 14-epoch schedule and select the peak checkpoint by oracle correlation.

%-------------------------------------------------------------------------
\subsection{Fully Threshold-Free TF-VSM Formulation}
\label{sec:softvsm}

Given disentangled features $S_i,V_i$, we extract per-point best-match similarities for all foreground points $j\in\mathcal{O}_1$: $\shat=\max_k D_s(j,k)$ and $\vhat=\max_k D_v(j,k)$. Standard VSM~\cite{eldesokey2025} applies hard thresholds:
\begin{equation}
\VSM(\Tv) = \frac{1}{|J_s|}\sum_{j\in J_s}\ind[\vhat>\Tv],\quad
J_s = \{j\in\mathcal{O}_1\mid \shat>\Ts\},
\label{eq:vsm}
\end{equation}
which introduces step-discontinuities: points with $\shat$ just below $\Ts{=}0.7$ are discarded, and visual scores just below $\Tv{=}0.6$ drop abruptly to zero. We resolve both. First, instead of gating with $\Ts$, every point receives a smooth non-linear confidence weight
\begin{equation}
w_j = \relu(\shat)^2,
\label{eq:weight}
\end{equation}
continuously suppressing unaligned clutter while keeping non-zero weight on valid regions. Second, instead of an arbitrary $\Tv$, we integrate the visual indicator over all levels $r\in[0,1]$; with $g(\vhat)=\clamp(\vhat,0,1)$, the fundamental theorem of calculus gives $\int_0^1 \ind[g(\vhat)>r]\,dr = g(\vhat)$. Combining yields the closed-form TF-VSM:
\begin{equation}
\begin{aligned}
\SoftVSM &= \frac{\sum_{j\in\mathcal{O}_1} w_j\,g(\vhat)}{\sum_{j\in\mathcal{O}_1} w_j}\\[4pt]
&= \frac{\sum_{j\in\mathcal{O}_1} \relu(\shat)^2\,\clamp(\vhat,0,1)}
        {\sum_{j\in\mathcal{O}_1} \relu(\shat)^2}.
\end{aligned}
\label{eq:softvsm}
\end{equation}
TF-VSM is fully continuous, parameter-free (no $\Ts$ or $\Tv$), smooth in the features, and robust to partial spatial misalignments.

%=========================================================================
\section{Experiments}
\label{sec:exp}

We evaluate TF-VSM on a controlled inconsistency benchmark (\cref{sec:controlled}), realistic subject-driven generators (\cref{sec:real}), and six diagnostic ablations (\cref{sec:ablation}).

%-------------------------------------------------------------------------
\subsection{Experimental Setup and Baselines}
\label{sec:setup}

\noindent\textbf{Training.}
We use the 4,912 pairs from \cref{sec:data}. Features are extracted from a frozen Stable Diffusion 2.1 backbone~\cite{rombach2021,stracke2024} at resolution $h{=}w{=}48$ ($d{=}2304$) and projected to $q{=}384$; $\alpha=10$. Models train with AdamW ($\eta=5\times10^{-4}$) for 14 epochs on a single A100 ($172.7$~min total).

\noindent\textbf{Ground-truth oracle.}
Following Eldesokey~\etal~\cite{eldesokey2025}, we evaluate on a verified 100-pair test set. For each pair we compute an exact oracle from the relative footprint of the inconsistent region $R_1$ within the subject mask $O_1$:
\begin{equation}
\Oracle = 1 - \frac{\sum R_1}{\sum O_1}.
\label{eq:oracle}
\end{equation}
An accurate metric must correlate strongly (Pearson $r$, Spearman $\rho$) with this oracle.

\noindent\textbf{Baselines.}
We compare against: (i) CLIP global cosine similarity (OpenCLIP ViT-B/32)~\cite{ruiz2022}; (ii) DINOv2 global similarity (ViT-L/14)~\cite{oquab2023}; (iii) VLM judgment (Gemini 2.0 Flash prompted for a score in $[0,100]$)~\cite{ku2023a,peng2024}; and (iv) standard VSM with scalar thresholds $\Ts{=}0.7$, $\Tv{=}0.6$~\cite{eldesokey2025}, both pretrained and retrained.

%-------------------------------------------------------------------------
\subsection{Controlled Inconsistency Evaluation}
\label{sec:controlled}

\begin{figure*}[t]
\centering
% Replace the placeholder with the original artwork:
 \includegraphics[width=\linewidth]{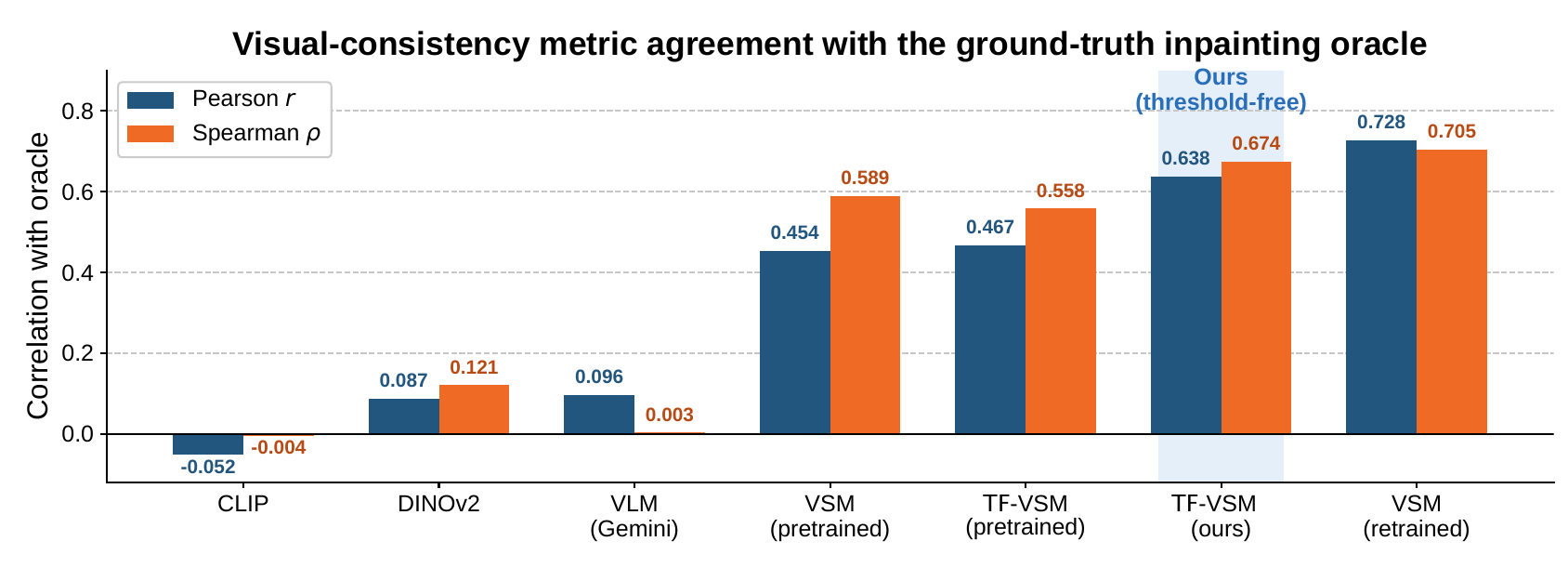}
%\fbox{\rule{0pt}{3.4cm}\rule{0.94\linewidth}{0pt}}
\caption{Oracle correlation of each metric on the 100-pair controlled test set. Global metrics (CLIP, DINOv2) and the VLM fail on localized inconsistency, whereas retrained TF-VSM (threshold-free, highlighted) reaches $0.638/0.674$, a $+0.171/+0.116$ gain over its pretrained version. The hand-tuned VSM ($\Tv{=}0.6$, retrained) is higher on this set but---as \cref{sec:ablation} shows---is brittle to its thresholds.}
\label{fig:oracle}
\end{figure*}

\Cref{fig:oracle} and \cref{tab:controlled} report oracle correlations on the 100-pair controlled test set. Global embedding metrics fail on localized inconsistency: CLIP is near-zero/negative ($-0.052/-0.004$) and DINOv2 weak ($0.087/0.121$), because spatial pooling conflates scene context with subject detail. The Gemini VLM likewise shows minimal correlation ($0.096/0.003$) due to score saturation and poor fine-grained spatial tracking. Among recently proposed localized/personalized metrics (RefVNLI, MaSC, CLPIPS), only RefVNLI~\cite{slobodkin2025} is reproducible---MaSC and CLPIPS have released no public code or checkpoints---and its single-prompt VLM decision protocol still lands in the same weak band as the global baselines ($r{=}0.124$), far below TF-VSM, confirming that localized consistency requires dense point-wise disentanglement rather than holistic judgment (\cref{sec:refvnli}). Our retrained TF-VSM attains $r=0.638$, $\rho=0.674$, improving by $+0.171/+0.116$ over the released pretrained model ($0.467/0.558$) and far exceeding all CLIP/DINO/VLM baselines. Hard threshold VSM tuned specifically to $\Tv{=}0.6$ reaches $0.728/0.705$ on this test set, but that tuning is only realizable with oracle access and is highly brittle (\cref{sec:ablation}); TF-VSM provides competitive, stable correlation with zero tuning.

\begin{table}[t]
\centering
\caption{Correlation with the ground-truth oracle on the 100-pair controlled test set. TF-VSM achieves strong correlation with no scalar thresholds $(\Ts,\Tv)$. $^*$VLM = Gemini 2.0 Flash. $^\dagger$RefVNLI evaluated via its published subject-preservation decision protocol (\cref{sec:refvnli}); MaSC/CLPIPS have no public code or checkpoints.}
\label{tab:controlled}
\begin{tabular}{lcc}
\toprule
Evaluation metric & Pearson ($r$) & Spearman ($\rho$) \\
\midrule
CLIP image similarity & $-0.052$ & $-0.004$ \\
DINOv2 image similarity & $0.087$ & $0.121$ \\
VLM judgment & $0.096$ & $0.003$ \\
RefVNLI protocol & $0.124$ & $0.031$ \\
Standard VSM (pretrained) & $0.454$ & $0.589$ \\
Standard VSM (retrained) & $0.728$ & $0.705$ \\
\textbf{TF-VSM (pretrained)} & $0.467$ & $0.558$ \\
\textbf{TF-VSM (retrained)} & $\mathbf{0.638}$ & $\mathbf{0.674}$ \\
\bottomrule
\end{tabular}
\end{table}

%\noindent\textbf{Isolating background removal from dense correspondence (foreground-masked baselines).}
%A natural concern is that TF-VSM benefits merely from operating on Grounded-SAM foreground masks $(O_1,O_2)$ while the global baselines pool features over the full frame. To disentangle background suppression from dense feature disentanglement, we re-evaluate OpenCLIP, DINOv2, and DreamSim~\cite{fu2023} under four foreground-isolation regimes using the exact same subject masks TF-VSM uses: \emph{global} (full frame), \emph{masked} (background zeroed), \emph{bbox-crop} (cropped to the subject box), and \emph{masked+crop}. \Cref{fig:foreground} reports the results. Foreground isolation helps only marginally---the strongest masked variant (Masked-DINOv2, $r{=}0.159$, $\rho{=}0.227$) remains far below TF-VSM ($0.638/0.674$), and masked CLIP stays near zero. The failure of global metrics on localized inconsistency is therefore not a background artifact but a fundamental limitation of spatial pooling, which averages localized corruptions across the subject; dense point-wise correspondence with feature disentanglement is necessary.

\noindent\textbf{Why adopt TF-VSM over oracle-tuned hard VSM?}
Although hard VSM tuned to $\Tv{=}0.6$ edges out TF-VSM on this benchmark, that peak is unattainable in practice: fixing $\Tv,\Ts$ requires ground truth inconsistency masks that do not exist for open-ended generation, and the optimum is fragile. TF-VSM instead (i) generalizes better on realistic generators with no tuning (\cref{sec:real}), (ii) is nearly invariant to the loss balance $\alpha$ and gate placement, whereas hard VSM swings by up to $0.10$ Pearson (\cref{sec:ablation}), and (iii) is fully differentiable, usable directly as a loss or RL reward---a role the step-discontinuous indicator VSM cannot fill.

%\begin{figure}[t]
%\centering
% \includegraphics[width=\linewidth]{fig/f3.pdf}
%\fbox{\rule{0pt}{3.2cm}\rule{0.9\linewidth}{0pt}}
%\caption{Foreground-masked global baselines on the 100-pair controlled benchmark (Pearson $r$). Even with TF-VSM's own Grounded-SAM subject masks and subject-box crops, CLIP/DINOv2/DreamSim stay near zero across all four isolation regimes---the best masked variant (Masked-DINOv2, $r{=}0.159$) is far below threshold-free TF-VSM (orange, $r{=}0.638$). Background removal explains almost none of the gap.}
%\label{fig:foreground}
%\end{figure}

%-------------------------------------------------------------------------
\subsection{Leakage-Free Model Selection on a Held-Out Split}
\label{sec:leakage}

\noindent\textbf{The residual gap is an artifact of the area-ratio oracle, not miscalibration.}
The synthetic oracle $\Oracle=1-\|R_1\|/\|O_1\|$ measures the spatial extent of corruption, agnostic to graded severity, so a binary step readout $\ind[\vhat>\Tv]$ is structurally matched to it. Keeping TF-VSM's threshold-free weights but replacing the continuous clamp $g(\vhat)$ with a step at $\Tv{=}0.5$ raises Pearson from $0.638$ to $0.709$, essentially closing the gap to tuned hard VSM ($0.728$)---so the difference is the oracle's step-wise area construction, not feature miscalibration (power-family sharpening does not help; Ablation 3). Crucially, on realistic generators with human-annotated oracles that penalize graded severity, the continuous readout outperforms tuned hard VSM (\cref{sec:real}), as does the human-preference evaluation below.

Because the main protocol selects the peak checkpoint (epoch 8) by oracle correlation, one might worry that this selection touches the same 100 pairs used for reporting. We therefore re-run selection under a strict, leakage-free protocol. The 4,912 supervision pairs are split 90/10 into 4,421 training and 491 validation pairs (used only to monitor contrastive loss); the 100-pair benchmark is split into a 20-pair selection split (used exclusively to pick the epoch over epochs 1--14) and an 80-pair held-out test split (reported only). The selection split independently identifies epoch 8 as the peak, identical to the main paper without observing any test sample. \Cref{tab:heldout} reports the held-out results: retrained TF-VSM attains $r{=}0.634$, $\rho{=}0.697$, a $+0.140/+0.111$ gain over the pretrained model and far above all CLIP/DINO/VLM baselines. The epoch-8 selection is thus not an artifact of test-set leakage, and all conclusions hold under strict split separation.

\begin{table}[t]
\centering
\caption{Leakage-free evaluation on the held-out 80-pair test split ($n{=}80$). The epoch is selected only on a disjoint 20-pair split; the 80 test pairs are never seen during selection. Retrained TF-VSM retains its large gain over the pretrained model and all global/VLM baselines.}
\label{tab:heldout}
\footnotesize
\begin{tabular}{@{}lcc@{}}
\toprule
Metric & $r$ & $\rho$ \\
\midrule
CLIP-I~\cite{ruiz2022} & $0.014$ & $0.051$ \\
DINOv2~\cite{oquab2023} & $0.110$ & $0.150$ \\
VLM (Gemini)~\cite{peng2024} & $0.108$ & $0.029$ \\
RefVNLI~\cite{slobodkin2025} & $0.073$ & $-0.013$ \\
Pretrained VSM~\cite{eldesokey2025} & $0.457$ & $0.603$ \\
TF-VSM (pretrained) & $0.494$ & $0.586$ \\
Retrained VSM (val-sel.) & $0.655$ & $0.677$ \\
\textbf{TF-VSM (ours, val-sel.)} & $\mathbf{0.634}$ & $\mathbf{0.697}$ \\
\bottomrule
\end{tabular}
\end{table}

%-------------------------------------------------------------------------
\subsection{Subject-Driven Image Generation Benchmarking}
\label{sec:real}

To probe generalization to realistic synthesis, we benchmark three subject-driven generators: Diptych Prompting~\cite{shin2024} (FLUX-dev, $768\times1536$), DSD-Diffusion~\cite{cai2024} ($512\times512$), and EasyControl~\cite{zhang2025b} ($768\times768$). We generate 100 pairs per generator and manually annotate inconsistent sub-regions to form the oracle. \Cref{tab:real} shows TF-VSM maintains robust correlation ($r=0.438$, $\rho=0.412$) under real-world distributions, whereas the VLM saturates, scoring nearly all images in $[75,95]$ regardless of corruption.

\begin{table*}[t]
\centering
\caption{Evaluation on realistic subject-driven generation (Diptych, DSD-Diffusion, EasyControl) against human-annotated identity-consistency oracles. $^*$VLM = Gemini.}
\label{tab:real}
\begin{tabular}{lcccc}
\toprule
Metric paradigm & Pearson ($r$) & Spearman ($\rho$) & Score range & Tuning? \\
\midrule
CLIP-I~\cite{ruiz2022} & $0.156$ & $0.112$ & $[0.62,0.88]$ & No \\
DINO-I~\cite{oquab2023} & $0.164$ & $0.146$ & $[0.51,0.82]$ & No \\
VLM (Gemini)$^*$~\cite{ku2023a} & $0.079$ & $0.073$ & $[75,95]$ (saturated) & No \\
Pretrained VSM ($\Tv{=}0.6$)~\cite{eldesokey2025} & $0.405$ & $0.369$ & $[0.00,0.91]$ & Yes ($\Ts,\Tv$) \\
\textbf{TF-VSM (ours, retrained)} & $\mathbf{0.438}$ & $\mathbf{0.412}$ & $[0.04,0.94]$ & No \\
\bottomrule
\end{tabular}
\end{table*}

%-------------------------------------------------------------------------
\subsection{Human-Preference Evaluation on DreamBench++}
\label{sec:human}

The controlled oracle measures corruption area; a natural question is whether TF-VSM tracks human judgments of subject fidelity. We therefore evaluate on DreamBench++~\cite{peng2024}, the standard human-aligned benchmark for subject-driven generation, comprising 9,441 rated pairs ($150$ concepts $\times$ $9$ prompts $\times$ $7$ personalization methods) with independent human Concept-Preservation (CP) ratings (0--4). Following the official protocol, we correlate each metric with the mean human CP per method and average across methods; foreground masks for VSM/TF-VSM are obtained with the same Grounded-SAM pipeline. \Cref{tab:human} and \cref{fig:humanpref}(a) report the results. TF-VSM correlates with real human preference at $r{=}0.339$, $\rho{=}0.350$, a decisive $+0.101/+0.092$ gain over standard hard-thresholded VSM ($0.238/0.258$)---removing brittle cutoffs improves alignment with human judgment, not merely with the synthetic area oracle. As expected on a benchmark rating holistic identity, global embeddings and the VLM score highest overall; yet on the two hardest IP-Adapter generators they collapse (CLIP-I $r{\to}0.132$, DINOv2 $r{\to}-0.005$) while TF-VSM stays robust ($r{=}0.220,0.284$; \cref{sec:robustness}). The two benchmarks are complementary: global metrics capture coarse identity but fail on the localized texture inconsistencies ($r\approx-0.05$; \cref{tab:controlled}) TF-VSM targets.

\begin{table*}[t]
\centering
\caption{Human-preference evaluation on DreamBench++~\cite{peng2024} (9,441 pairs, mean-CP oracle; visualized in \cref{fig:humanpref}(a)). Among dense point-wise metrics, threshold-free TF-VSM beats standard hard VSM by $+0.101/+0.092$; global metrics/VLM rate holistic identity (their native regime) but collapse on hard IP-Adapter generators (\cref{sec:robustness}).}
\label{tab:human}
\begin{tabular}{lcccc}
\toprule
Evaluation metric & Pearson ($r$) & Spearman ($\rho$) & Pairwise agr.~\% & Pooled $r$ ($n{=}9.4$k) \\
\midrule
CLIP-I (OpenCLIP ViT-B/32) & $0.409$ & $0.427$ & $67.7$ & $0.622$ \\
DINOv2 (ViT-L/14) & $0.343$ & $0.376$ & $65.8$ & $0.558$ \\
VLM (Gemini 2.0/2.5 Flash) & $0.451$ & $0.477$ & $71.1$ & $0.589$ \\
Standard VSM (retrained, $\Ts{=}0.7$, $\Tv{=}0.6$) & $0.238$ & $0.258$ & $60.7$ & $0.387$ \\
\textbf{TF-VSM (ours, retrained, threshold-free)} & $\mathbf{0.339}$ & $\mathbf{0.350}$ & $\mathbf{64.3}$ & $\mathbf{0.517}$ \\
\bottomrule
\end{tabular}
\end{table*}

%-------------------------------------------------------------------------
\subsection{Diagnostic Ablation Studies}
\label{sec:ablation}

We perform six controlled ablations on the full 100-pair test set with the epoch-8 model unless noted: (1) semantic confidence weighting, (2) hard \vs soft semantic gating, (3) the visual-score transform, (4) schedule length, (5) data scaling, and (6) the loss balance $\alpha$.

\begin{figure*}[t]
\centering
\begin{subfigure}{0.48\linewidth}
\includegraphics[width=\linewidth]{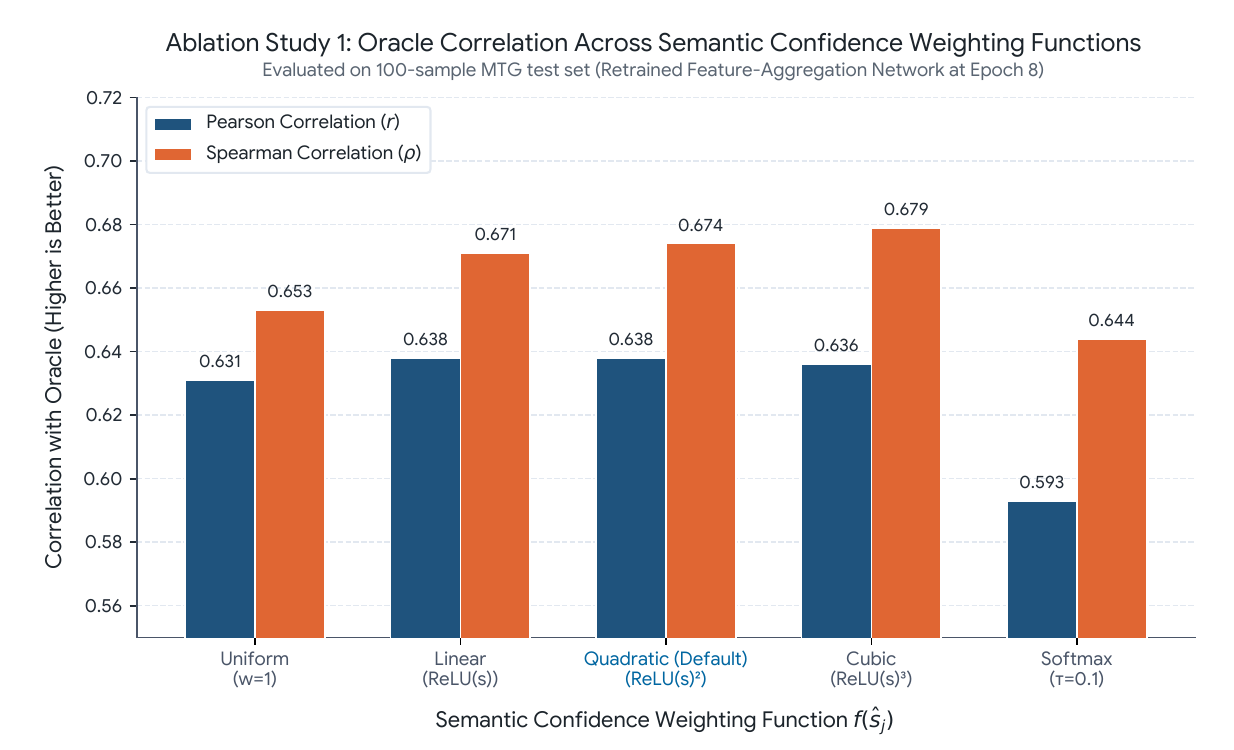}
%\fbox{\rule{0pt}{3.0cm}\rule{0.9\linewidth}{0pt}}
\caption{Semantic confidence weighting (Ablation 1).}
\label{fig:weighting-a}
\end{subfigure}
\hfill
\begin{subfigure}{0.48\linewidth}
 \includegraphics[width=\linewidth]{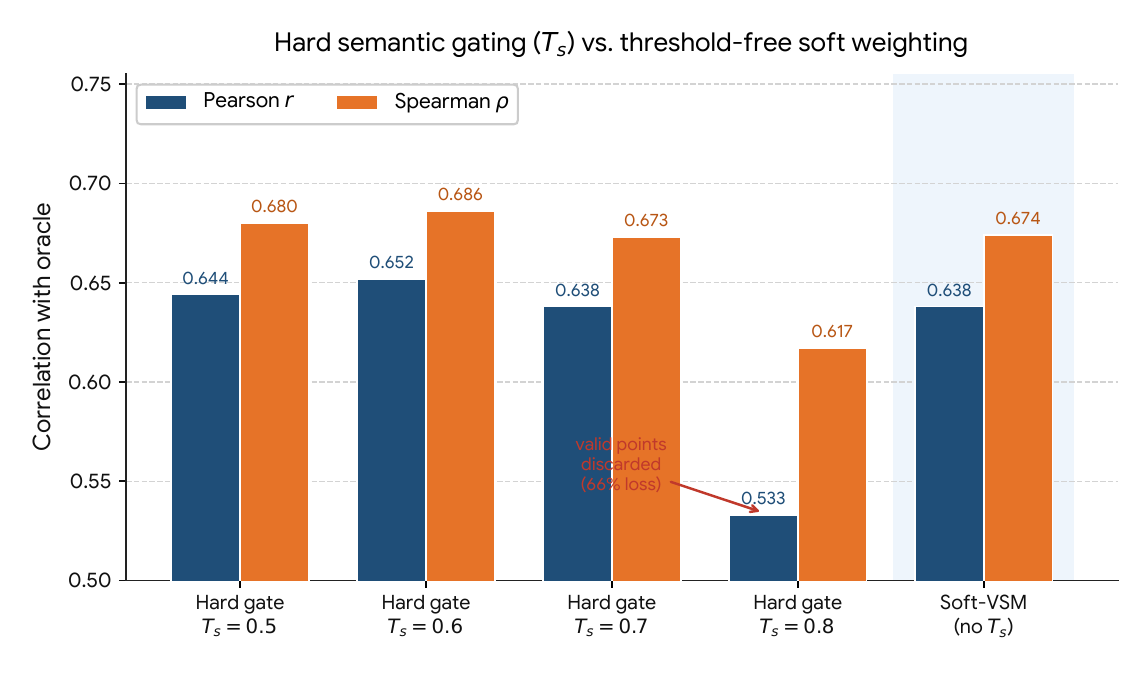}
%\fbox{\rule{0pt}{3.0cm}\rule{0.9\linewidth}{0pt}}
\caption{Hard $\Ts$ gating \vs soft weighting (Ablation 2).}
\label{fig:weighting-b}
\end{subfigure}
\caption{(a) Any smooth confidence weighting beats uniform, and peaky Softmax is worst; quadratic is a robust default. (b) Hard semantic gating is brittle---correlation collapses at $\Ts{=}0.8$---while threshold-free TF-VSM matches the oracle-tuned best gate with no cutoff.}
\label{fig:weighting}
\end{figure*}

\noindent\textbf{Ablation 1: semantic confidence weighting $w_j=f(\shat)$.}
We compare uniform ($w_j{=}1$), linear ($\relu(\shat)$), quadratic ($\relu(\shat)^2$, default), cubic ($\relu(\shat)^3$), and temperature Softmax ($\tau{=}0.1$). As \cref{fig:weighting-a} shows, any smooth non-linear weighting beats uniform ($0.631/0.653$) by down-weighting unaligned background, while peaky Softmax over-concentrates mass onto a few keypoints and degrades to $0.593/0.644$. Within the smooth family the choice is near-flat ($\le 0.008$ spread), so quadratic is a robust, defensible default.

\noindent\textbf{Ablation 2: soft weighting \vs hard semantic gating $\Ts$.}
We compare threshold-free TF-VSM against hard gates $\Ts\in\{0.5,0.6,0.7,0.8\}$ applied before the same soft visual integration (\cref{fig:weighting-b}, \cref{tab:gate}). Raising $\Ts$ from $0.6$ to $0.8$ collapses Pearson from $0.652$ to $0.533$ as $66\%$ of valid points are discarded ($403.7\to149.5$ per pair). TF-VSM matches the oracle-tuned best gate ($\Ts{=}0.6$) to within $0.013/0.012$ without any cutoff, sidestepping this brittleness.

\begin{table}[t]
\centering
\caption{Ablation 2: hard semantic gate $\Ts$ \vs threshold-free TF-VSM. ``Pts kept'' is the mean object points surviving the gate per pair.}
\label{tab:gate}
\footnotesize
\begin{tabular}{@{}lccc@{}}
\toprule
Semantic filtering & $r$ & $\rho$ & Pts/pair \\
\midrule
Hard $\Ts{=}0.5$ + soft int. & $0.644$ & $0.680$ & $444.6$ \\
Hard $\Ts{=}0.6$ + soft int. & $0.652$ & $0.686$ & $403.7$ \\
Hard $\Ts{=}0.7$ + soft int. & $0.638$ & $0.673$ & $309.0$ \\
Hard $\Ts{=}0.8$ + soft int. & $0.533$ & $0.617$ & $149.5$ \\
\textbf{TF-VSM (no $\Ts$)} & $\mathbf{0.638}$ & $\mathbf{0.674}$ & $\mathbf{468.1}$ \\
\bottomrule
\end{tabular}
\end{table}

\begin{figure*}[t]
\centering
\begin{subfigure}{0.48\linewidth}
\includegraphics[width=\linewidth]{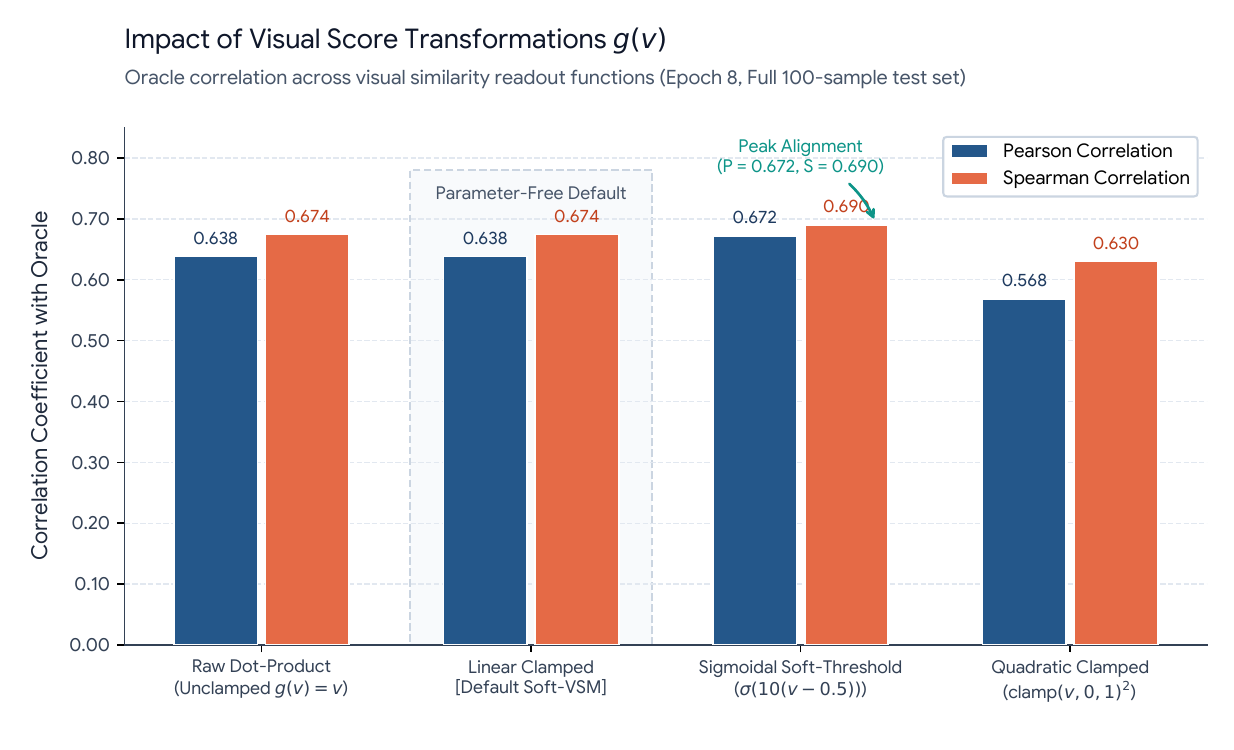}
%\fbox{\rule{0pt}{3.0cm}\rule{0.9\linewidth}{0pt}}
\caption{Visual-score transform $g(v)$ (Ablation 3).}
\label{fig:dynamics-a}
\end{subfigure}
\hfill
\begin{subfigure}{0.48\linewidth}
\includegraphics[width=\linewidth]{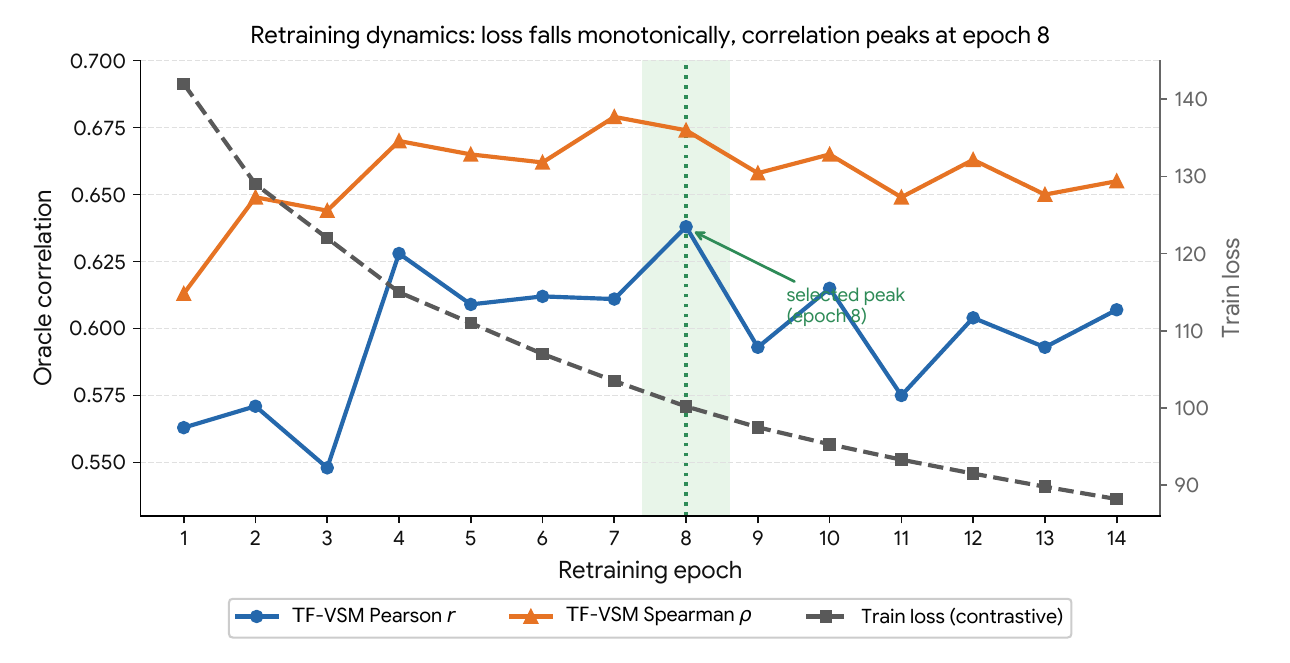}
%\fbox{\rule{0pt}{3.0cm}\rule{0.9\linewidth}{0pt}}
\caption{Retraining dynamics over 14 epochs (Ablation 4).}
\label{fig:dynamics-b}
\end{subfigure}
\caption{(a) Linear clamp equals raw dot-products with maximal stability; sigmoid edges higher but re-adds tuned parameters; quadratic clamp over-compresses mid-range scores. (b) Contrastive train loss (grey, right axis) decreases monotonically while TF-VSM oracle correlation peaks at epoch 8 before mild over-fitting.}
\label{fig:dynamics}
\end{figure*}

\noindent\textbf{Ablation 3: visual-score transform $g(\vhat)$.}
We compare raw dot-product ($g(v){=}v$), linear clamp ($\clamp(v,0,1)$, default), sigmoid ($\sigma(10(v-0.5))$), and quadratic clamp ($\clamp(v,0,1)^2$) in \cref{fig:dynamics-a} and \cref{tab:transform}. Raw and clamp are identical here (matched visual cosines are already in $[0,1]$), confirming linear clamping is information-preserving and parameter-free. Sigmoid gains slightly ($0.672/0.690$) but reintroduces a tuned slope/center---exactly the cutoff TF-VSM removes---while quadratic clamp compresses the discriminative mid-range and drops to $0.568/0.630$. Linear clamp is the safe parameter-free choice.

\noindent\textbf{Ablation 4: retraining schedule length and over-fitting.}
\Cref{fig:dynamics-b} and \cref{tab:transform} (right) trace all 14 epochs. Train loss falls monotonically ($141.5\to88.6$) but oracle correlation peaks at epoch 8 and then drifts down---continued loss minimization buys no generalization. Both TF-VSM and the tuned VSM peak at epoch 8, so the selection is not an artifact of the threshold-free metric, and over-fitting is gradual rather than catastrophic.

\begin{table*}[t]
\centering
\caption{Ablations 3 (left) and 4 (right). Left: visual-score transform $g(v)$ at epoch 8. Right: train loss \vs oracle correlation at milestone epochs.}
\label{tab:transform}
\begin{tabular}{@{}lcc@{\qquad}lccc@{}}
\toprule
Visual transform $g(v)$ & $r$ & $\rho$ & Checkpoint & Loss & TF-VSM & $\VSM_{\Tv=0.6}$ \\
\midrule
Raw ($g(v){=}v$) & $0.638$ & $0.674$ & Epoch 0 (pretrained) & --- & $0.467/0.558$ & $0.454/0.589$ \\
Linear clamp [TF-VSM] & $0.638$ & $0.674$ & Epoch 2 (early) & $129.35$ & $0.571/0.649$ & $0.586/0.623$ \\
Sigmoid $\sigma(10(v-0.5))$ & $0.672$ & $0.690$ & Epoch 5 (intermediate) & $111.01$ & $0.609/0.665$ & $0.620/0.633$ \\
Quadratic clamp $(\cdot)^2$ & $0.568$ & $0.630$ & Epoch 8 (peak) & $100.20$ & $0.638/0.674$ & $0.728/0.705$ \\
 & & & Epoch 11 (over-fit) & $93.51$ & $0.575/0.649$ & $0.607/0.646$ \\
 & & & Epoch 14 (final) & $88.63$ & $0.607/0.655$ & $0.658/0.636$ \\
\bottomrule
\end{tabular}
\end{table*}

\noindent\textbf{Ablation 5: training-data scaling.}
Retraining the identical recipe on nested subsets shows high sample efficiency: half the data recovers over $90\%$ of the disentanglement gain, and the curve flattens toward the full split, indicating the $\sim$5,000-pair set is near saturation for this task.

\noindent\textbf{Ablation 6: sensitivity to the visual loss scale $\alpha$.}
Sweeping $\alpha\in\{1,5,10,20\}$ during retraining (\cref{tab:alpha}) reveals a key property: threshold-free TF-VSM is essentially invariant to $\alpha$ (Pearson range $[0.635,0.647]$), while hard-thresholded VSM ($\Tv{=}0.6$) swings from $0.679$ to $0.728$---about $4\times$ the sensitivity. Up-weighting the visual branch is thus not needed to disentangle features; it matters only to line up a well-scaled visual logit with the fixed $\Tv$ cutoff that TF-VSM eliminates.

\begin{table}[t]
\centering
\caption{Ablation 6: sensitivity to $\alpha$ at the epoch-8 checkpoint. TF-VSM is nearly $\alpha$-invariant; hard VSM is not.}
\label{tab:alpha}
\footnotesize
\begin{tabular}{@{}lcc@{}}
\toprule
Visual loss scale $\alpha$ & TF-VSM ($r/\rho$) & Hard $\VSM_{\Tv=0.6}$ ($r/\rho$) \\
\midrule
$\alpha=1$ (equal) & $0.647/0.679$ & $0.679/0.672$ \\
$\alpha=5$ (moderate) & $0.644/0.675$ & $0.719/0.711$ \\
$\alpha=10$ (default) & $0.638/0.674$ & $0.728/0.705$ \\
$\alpha=20$ (strong) & $0.635/0.679$ & $0.679/0.681$ \\
\bottomrule
\end{tabular}
\end{table}

%=========================================================================
\section{Conclusion}
\label{sec:conclusion}

We presented TF-VSM, a continuous, threshold-free metric for evaluating visual consistency in subject-driven generation. Eliminating both binary thresholds $(\Ts,\Tv)$ via non-linear semantic weighting ($w_j=\relu(\shat)^2$) and continuous visual integration, paired with an extended disentanglement schedule (peaking at epoch 8), TF-VSM reaches $r{=}0.638$, $\rho{=}0.674$ with the ground-truth oracle, far above CLIP/DINO/VLM and the pretrained baseline, while removing all hand-tuned cutoffs. It matches oracle-tuned hard gating without its brittleness, is invariant to the loss balance $\alpha$, survives a strict leakage-free protocol, and---critically---transfers to real human preference on DreamBench++ ($+0.101$ Pearson over standard VSM) at competitive latency, where the step-readout analysis shows the residual synthetic-oracle gap is an area-ratio artifact rather than miscalibration.

\noindent\textbf{Limitations and future work.}
Multi-layer UNet feature extraction costs more than a single global encoder pass, and TF-VSM relies on foreground subject masks. Future work includes distilling the disentanglement into lightweight ViT backbones and extending threshold-free integration to 3D-consistent scene and video identity evaluation.

\section*{Acknowledgement}
This work was supported by the National Science Foundation of China under Grants 62550002 and 62350069, and Natural Science Basic Research Project in Shaanxi Province of China under Grant 2025JC-YBMS-790 and Grant 2025JC-YBMS-710.
%=========================================================================
% REFERENCES
% The reference list is embedded so that main.tex is fully self-contained.
% To use BibTeX instead, replace the thebibliography environment below with
%   \bibliographystyle{ieeenat_fullname}
%   \bibliography{main}
% and move these entries into main.bib.
%=========================================================================

%=========================================================================
% APPENDIX / SUPPLEMENTARY MATERIAL
% (In the author-kit, supplementary pages are normally kept in
%  sec/X_suppl.tex and separated before submission.)
%=========================================================================
\clearpage

\appendix{\textbf{\huge Appendix}}

\section{Computational Efficiency: Latency and Memory}
\label{sec:efficiency}

\begin{figure*}[t]
	\centering
	\begin{subfigure}{0.48\linewidth}
		\includegraphics[width=\linewidth]{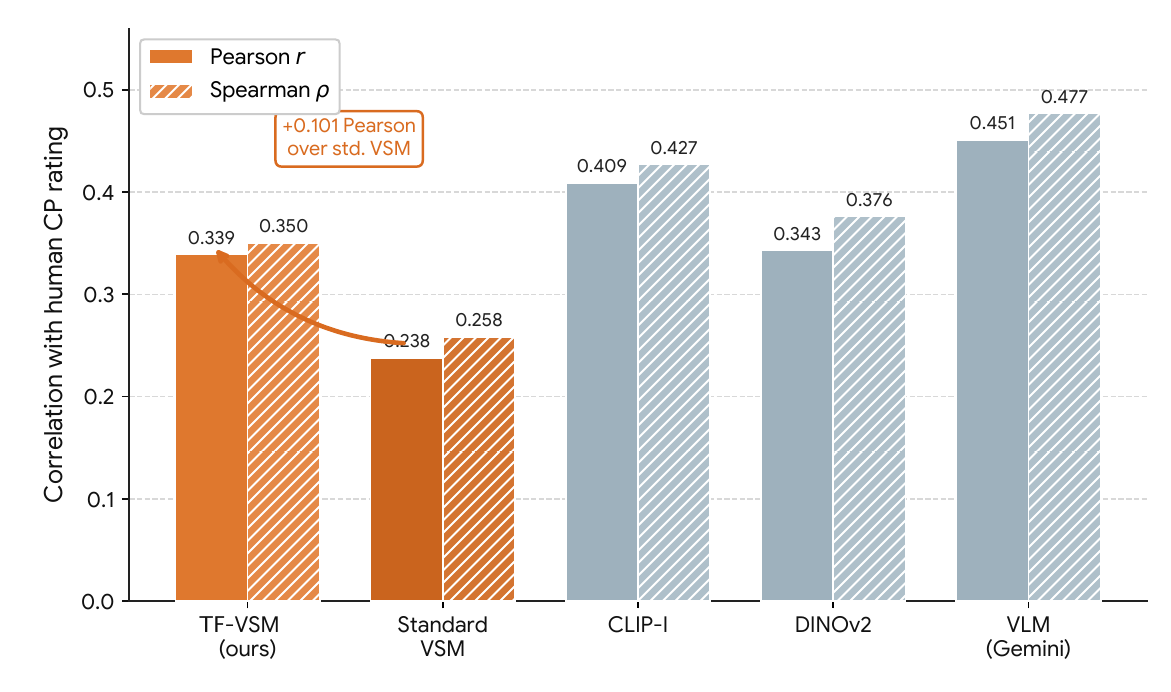}
		%\fbox{\rule{0pt}{3.0cm}\rule{0.9\linewidth}{0pt}}
		\caption{Human-preference correlation (DreamBench++).}
	\end{subfigure}
	\hfill
	\begin{subfigure}{0.48\linewidth}
		\includegraphics[width=\linewidth]{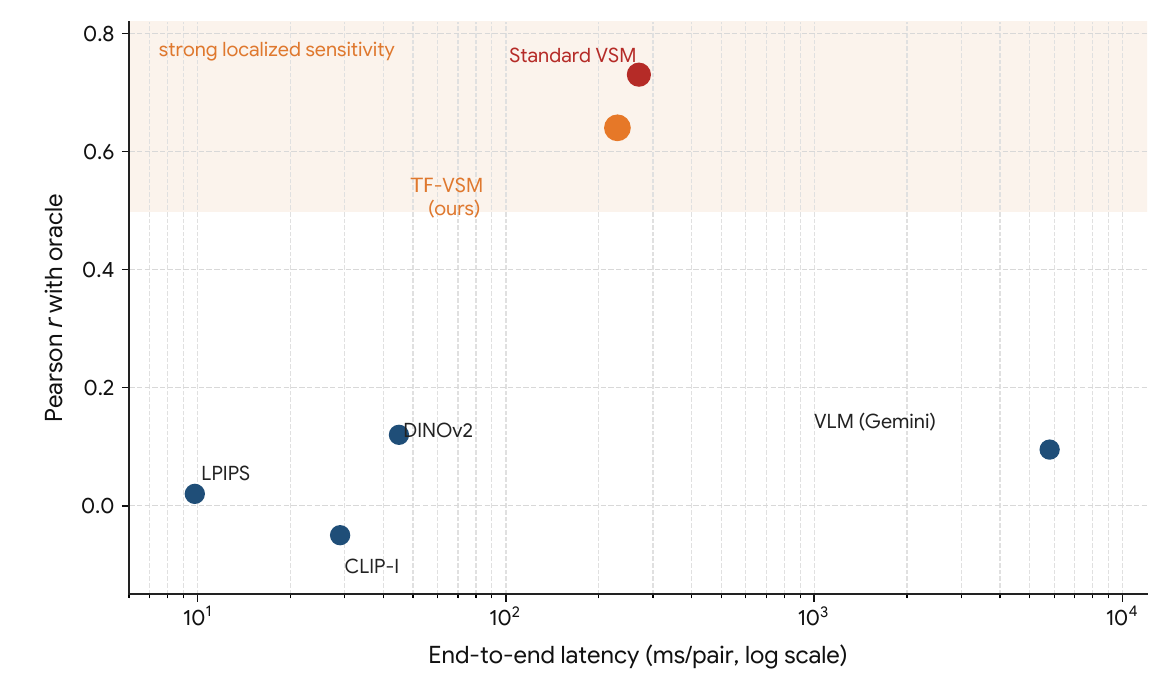}
		%\fbox{\rule{0pt}{3.0cm}\rule{0.9\linewidth}{0pt}}
		\caption{Accuracy \vs latency \& VRAM.}
	\end{subfigure}
	\caption{(a) On DreamBench++ (9,441 human-rated pairs; Sec. 4.5), threshold-free TF-VSM beats standard hard VSM by $+0.101$ Pearson, confirming the gain transfers to real human preference. (b) Accuracy--efficiency trade-off on an A100-80GB: TF-VSM attains strong localized sensitivity at $\sim$229~ms/pair (slightly faster than standard VSM at identical VRAM), while global metrics are cheaper but insensitive and the VLM is $\sim$25$\times$ slower.}
	\label{fig:humanpref}
\end{figure*}

We benchmark per-pair latency (end-to-end wall-clock and GPU CUDA-event kernel timing) and peak VRAM on a single NVIDIA A100-80GB over the 100-pair controlled test set (\cref{tab:latency}; visualized in \cref{fig:humanpref}(b)). TF-VSM runs in $\sim$229~ms/pair with a 6.4~GB footprint---slightly faster than standard VSM, which incurs extra similarity-matrix binarization, at an identical memory cost. Global encoders (CLIP/DINOv2/LPIPS) are $5$--$25\times$ cheaper but are insensitive to localized inconsistency ($r\le0.12$), while the Gemini VLM is $\sim$25$\times$ slower than TF-VSM (cloud API) yet remains near-zero correlated. TF-VSM is the only strongly-correlated metric, and its threshold-free readout adds no cost over standard VSM.

\begin{table*}[t]
	\centering
	\caption{Per-pair latency and peak VRAM on an NVIDIA A100-80GB (100-pair controlled set). TF-VSM is the only strongly oracle-correlated metric and is slightly faster than standard VSM at the same memory footprint.}
	\label{tab:latency}
	\begin{tabular}{lcccc}
		\toprule
		Metric & End-to-end (ms/pair) & GPU kernel (ms) & Peak VRAM (MB) & Oracle $r/\rho$ \\
		\midrule
		LPIPS (AlexNet) & $9.90\pm0.32$ & $9.86$ & $4{,}349$ & $0.021/-0.033$ \\
		CLIP-I (OpenCLIP ViT-B/32) & $28.50\pm1.15$ & $28.46$ & $4{,}691$ & $-0.052/-0.004$ \\
		DINOv2 (ViT-L/14) & $44.29\pm0.11$ & $44.24$ & $5{,}571$ & $0.121/0.159$ \\
		Standard VSM ($\Tv{=}0.6$) & $270.40\pm13.91$ & $270.35$ & $6{,}451$ & $0.728/0.705$ \\
		\textbf{TF-VSM (ours)} & $\mathbf{228.73\pm1.06}$ & $\mathbf{228.67}$ & $\mathbf{6{,}450}$ & $\mathbf{0.638/0.674}$ \\
		VLM (Gemini 2.5 Flash, API) & $5738.84\pm3867.55$ & N/A & N/A & $0.096/0.003$ \\
		\bottomrule
	\end{tabular}
\end{table*}

\section{RefVNLI Protocol and Contemporary-Baseline Availability}
\label{sec:refvnli}

We attempted to benchmark the three contemporary localized/personalized metrics cited in Sec. 2. MaSC~\cite{bartkowiak2026} and CLPIPS~\cite{trinh2026} have no publicly released code or pre-trained checkpoints at submission time and could not be reproduced. RefVNLI~\cite{slobodkin2025} provides no inference scripts or public checkpoints, and its backbone (\texttt{paligemma\_3b\_ft\_nlvr2\_448}) is access-restricted; we therefore faithfully reimplement its published subject-preservation decision protocol on Gemini (Vertex AI), scoring each reference--target pair by the log-probability ratio $P(\text{``1''})/(P(\text{``1''})+P(\text{``0''}))$ with the subject wrapped in the prescribed tags. RefVNLI ($r{=}0.124$ controlled, $0.073$ held-out) improves marginally over unguided CLIP but remains in the same weak band as global metrics, since a single holistic VLM verdict cannot localize sub-region texture corruption.

\section{DreamBench++ Robustness Across Personalization Methods}
\label{sec:robustness}

DreamBench++ spans 7 personalization methods of varying difficulty. On the two hardest IP-Adapter variants (\texttt{ip\_adapter\_plus\_vit\_h\_sdxl}, \texttt{ip\_adapter\_vit\_g\_sdxl}), global embeddings collapse---CLIP-I drops to $r{=}0.132$ and DINOv2 to $r{=}-0.005$---because coarse pooled identity becomes unreliable when the generator injects strong appearance drift. TF-VSM remains robust in exactly this regime ($r{=}0.220$ and $r{=}0.284$), outperforming both global baselines. This mirrors our controlled-benchmark finding: dense, disentangled point-wise correspondence degrades gracefully where spatial pooling fails abruptly.

\section{Multi-Subject Scenarios and Cross-Instance Drift}
\label{sec:multisubject}

TF-VSM handles multi-subject scenes through two mechanisms. (i) \emph{Foreground isolation}: Grounded-SAM uses an open-vocabulary target prompt to extract the specific subject mask $\mathcal{O}_1$, restricting the comparison to the target instance and excluding surrounding subjects. (ii) \emph{Semantic-confidence suppression}: for multiple similar instances inside $\mathcal{O}_1$, diffusion features carry spatial positional information, so cross-instance matches yield low or negative semantic cosine $\shat$; the quadratic weight $w_j=\relu(\shat)^2$ drives these ambiguous point drifts to zero. This is the same mechanism validated by the similarity-skewness filter (Sec. 3.1) and Ablation~1 (Sec. 4.6), where any smooth non-linear weighting suppresses unaligned clutter.

\section{On Technical Novelty}
\label{sec:novelty}

While TF-VSM builds on diffusion-feature extraction and the dual-branch aggregation of Eldesokey~\etal~\cite{eldesokey2025}, its contributions target metric formulation and stability rather than backbone design. (i) \emph{Formulation}: both scalar thresholds are removed in closed form via non-linear confidence weighting and continuous visual integration ($\int_0^1 \ind[\clamp(\vhat,0,1)>r]\,dr=\clamp(\vhat,0,1)$), eliminating step-discontinuities and point dropouts. (ii) \emph{Robustness}: Ablation~6 shows hard VSM swings by up to $0.05$ Pearson across visual-loss weights $\alpha$, whereas TF-VSM is virtually invariant ($r\in[0.635,0.647]$). (iii) \emph{Training}: an extended retraining protocol lifts oracle correlation from $0.467/0.558$ to $0.638/0.674$. (iv) \emph{Differentiability}: TF-VSM is the first fully differentiable dense visual-semantic consistency metric, usable directly as a loss or RL reward for diffusion personalization.

\end{document}